# Facial Information Analysis Technology for Gender and Age Estimation

Gilheum Park
KyungPook National University
80, Daehak-ro, Buk-gu, Daegu, Republic of Korea
khpark@ee.knu.ac.kr

Sua Jung
University of Massachusetts Lowell
220 Pawtucket St, Lowell, MA 01854, USA
sua_jung@student.uml.edu

## Abstract

*This is a study on facial information analysis technology for estimating gender and age, and poses are estimated using a transformation relationship matrix between the camera coordinate system and the world coordinate system for estimating the pose of a face image. Gender classification was relatively simple compared to age estimation, and age estimation was made possible using deep learning-based facial recognition technology. A comparative CNN was proposed to calculate the experimental results using the purchased database and the public database, and deep learning-based gender classification and age estimation performed at a significant level and was more robust to environmental changes compared to the existing machine learning techniques.*

## 1. Introduction

This paper describes research on facial information analysis technology for age and gender estimation. In this study, age estimation and gender classification were done using face images, and to estimate the pose of the face image, the pose is evaluated using a transformation relation matrix between the camera coordinate system and the world coordinate system. In addition, EigenFace and FisherFace were used to investigate the face recognition algorithm using the existing machine learning technique. The results prove that age estimation is more difficult than gender classification. Because the biological age of the face is different from the actual age of the subject, estimating the age is a very difficult problem. It was confirmed that the solution using deep learning-based facial recognition technology is possible. We propose a comparative CNN to calculate the experimental results using an actual public database or a purchased database. Deep learning-based gender classification and age estimation showed significant performance compared to existing machine learning techniques and was resistant to environmental changes. The proposed technology can be used in various purpose, for example, personal authorization and marketing if a face is searched fast by using this technology.

## 2. Related Works

### 2.1 Existing Research

**Eigen Face** EigenFace is selected as a representative study in the field of face recognition, and after extracting the feature points, PCA (Principal Component Analysis) is performed on the feature points, and the Euclidean distance between the recognized face target and the correct answer is applied to recognize it. Although it responds sensitively to changes in lighting and surroundings, it is widely used and a proven representative method in the field of face recognition.

**FisherFace** It is a face recognition method using FLD (Fisher Linear Discriminant) as a classification algorithm. This is a method to find a coefficient that can distinguish different groups well when data is expressed as a linear combination of young variables. It is a method used in real-time face recognition because it is more accurate by learning the individual characteristics of the face and has characteristics that are not sensitive to environmental changes compared to EigenFace.

### 2.2 CNN-based Gender Classification

**AlexNet** This is a thesis by Supervision team at the University of Toronto, Canada, who won the 2012 ILSVRC (ImageNet Large Scale Visual Recognition Challenge) based on ImageNet DB. Compared to the performance of other participants' algorithms, it took the first place with overwhelming performance, and once again became an opportunity to actively conduct research on deep learning with MLP (Multi-Layer Perceptron) as its parent. AlexNet consists of 5 convolution layers and 3 fully connected layers.

**Inception** Inception shows a very complex structure compared to the structure of AlexNet, ZFNet, and LeNet, and the depth of the network is deepened rapidly. The problem that arises as the network deepens is that as the number of parameters increases, the possibility of overfitting the DB increases, and the amount of computation increases. To solve this problem, a study called Inception Module was proposed. Efforts were made to strengthen the scale variation by applying convolution with different twists. By properly using 1x1 convolution, it has the effect of reducing the dimension and solving the problem of increasing the amount of computation when the network is deep.

**Simpler Model (Modified AlexNet)** In the Simpler Model, by modifying the existing AlexNet, four convolution layers were removed, and one convolution layer with a size of 3x3 was used and three fully connected layers were used. The experimental results are confirmed in the next section to calculate the performance when this is applied to a small model rather than a model with a deep network depth.

**Teacher-Student Model** The Teacher-Student Model is a research field that is conducted to make the network model of deep learning more compact, and it is a method that can simplify the model structure while showing performance comparable to the existing deep network models. Various studies have been conducted to better convey knowledge, and FitNet assigns a guided layer to the hidden layer to better perform distilling the knowledge and defines a loss function so that the specified layer is imitated by the student layer to effectively convey knowledge.

**MS Face API** MS Face API is the Face Recognition API provided by Microsoft and provides APIs in various languages to develop applications using the Perceptual Intelligence function. It is possible to check the excellent performance with 96% of the result of performing the DB targeting.

## 3. Problem Formulation

Efficient routes for collecting moving object (person) information are very diverse. In addition to security cameras for methods, various IoT (Internet of Things) products due to the 4th industrial revolution have appeared in the market. From various domestic institutions to households, the installation of cameras has become common, and a vast amount of object information is being collected through this. However, the establishment of a system for monitoring the information acquired through the installed camera is insufficient. To extract specific object information in case of emergency, it is very inefficient because it must be done manually from a vast number of stored images. To overcome this, we propose the results of research on a high-speed face information estimation technology that can automatically extract the information of a moving object from an image. It is a technology that makes it possible to extract information for a specific purpose by studying high-speed face information estimation technology that can automatically extract information of moving objects.

## 4. Methods

### 4.1 Age Estimation

**DEX (Deep EXception)** DEX overcomes the limitation of insufficient DB in estimating the apparent age by applying a fine-tuning technique labeled by age showing a CNN model trained with a DB labeled with real age. Also, in regressing the age using the classification technique, the age was estimated by calculating the expected value by looking at the output of each class of the classifier as the probability of belonging to the class.

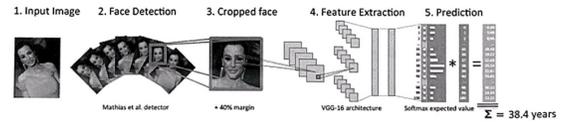

Figure 1. DEX Technique Flow Chart

After detecting the ROI corresponding to the face region by rotating the input image at various angles, the ROI with the highest score is selected for the detected ROIs, and the image is cropped with a 40% margin in the up, down, left, and right directions. The goal of this process is to align the face image by correcting the rotational transformation. This is a method of estimating the age of the input image by using the sorted image as an input to the VGG-16 architecture, looking at the value of the SoftMax layer, which is the result of the architecture, as the probability of belonging to each class, and obtaining the expected value. Mean Absolute Error (MAE) is the average value of the absolute value of the error of the estimate and the correct answer and is a protocol for validation generally used to measure the result of age estimation. The formula is: n is the number of tests DBs, $y_i$ is the estimate, and $x_i$ is the correct answer.

$$MAE = \frac{\sum_{i=0}^{n}|y_i - x_i|}{n}$$

Formula 1. MAE Calculation Formula

**Ranking CNN for Age Estimation** The Ranking CNN for Age Estimation technique uses binary CNNs as many as the number of classes to determine whether a specific class is older or younger and ranks the output of each binary

CNN for one input image to finally determine the age of the input image. estimated. The corresponding binary CNN outputs 1 if the input image is younger than the age of the current class, and 0 if it is more. This can be expressed as a formula as follows. $D_k$ is the kth class binary CNN output and $y_i$ is the label of the input image. To determine the age of the input image using the binary CNN configured as above, the Ranking CNN for Age Estimation method uses as many binary CNNs as the number of classes.

$$D_x = 1 \ (y_i < k) \ or \ 0 \ (y_i \geq k)$$
Formula 2. Binary CNN Output

**Comparative Deep Learning** In Comparative Deep Learning method, one of the input images in Chapter 2 is fixed as the base line, and all images in the train DB are variably input in the other one and compared with the baseline. When two input images are input to the CNN, the energy is measured by calculating the distance between the last fully connected layers of each CNN model. By constructing a loss function based on the measured energy, the distance between fully connected layers decreases when the age of the input image is younger than the baseline image, and the distance between fully connected layers increases when the age of the input image is older than the baseline image. To finally estimate the age of the input image from the binary CNN model that judges whether the input image is older than the baseline image, Comparative Deep Learning uses several binary CNNs.

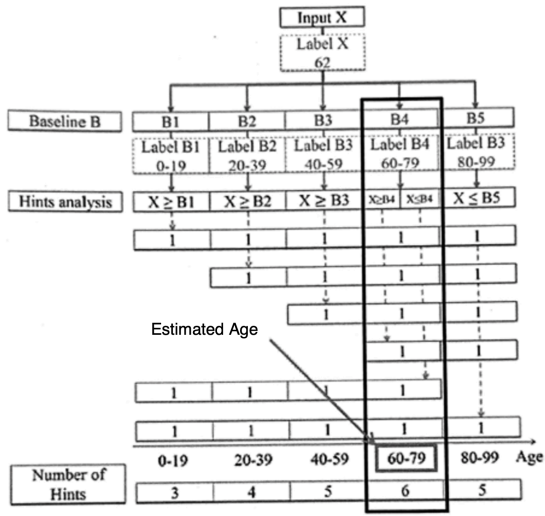

Figure 2. Age estimation method from 5 Binary CNNs

In Figure 2, B1 to B5 mean the constructed binary CNNs. If the input image of each binary CNN is estimated to be older than the baseline, the older class is all 1, and conversely, if the age is estimated to be small, the youngest class is all 1. to constitute Hints analysis. After that, by summing hits for each class, the class with the maximum sum of hints is finally estimated as the age of the input image.

**Age Estimation using Comparative CNN**

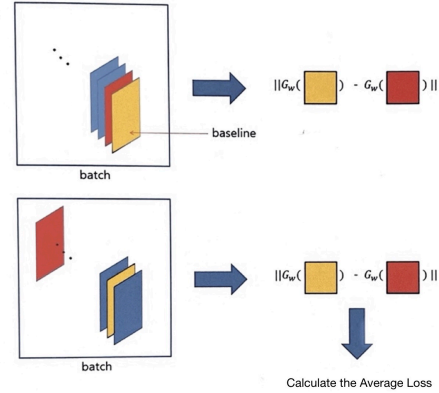

Figure 3. Comparative CNN Technique Implementation

In implementing Comparative CNN, the method shown in Figure 3 was applied. View a specific image as a baseline in one batch, calculate each energy with the rest of the images, and substitute the loss function below to find the average of the loss, change the baseline, and find the average of the obtained loss averages to learn in the direction of minimizing the loss. The Loss function can be arranged as follows.

$$L = (1 - Z_i) \ L^- \ (E \ (X_i, B_m)) + (Z_i) \ L^+ \ (E \ (X_i, B_m)).$$
$Z_i$: same class goes to 1, another class goes to 0
$L^-$: Decrease Function
$L^+$: Increase Function
$E( )$: Distance Function
X, B: input image

Formula 3. The Loss Function

### 4.2 Sex and Age Estimation

**Gender and Age Estimation using Comparative CNN-Based Multi-task Learning** Gender and Age Estimation Method Using Comparative CNN-Based Multi-task Learning In estimating age using Comparative CNN, a multi-task method of estimating gender was applied. In Comparative CNN, the number of dimensions of the (last fully connected layer) is set to 70, and the feature vector for gender classification is set to 10 dimensions. function was used, and the comparative CNN loss function for classifying gender was used for the 10-dimensional fully connected layer. The rest of the implementation method is the same as the age estimation method using Comparative CNN.

## 5. Experiments

### 5.1 Performance and Comparison

Figure 4 shows the experimental results applied to Mega Asian, an open face image DB, using the implemented Comparative CNN. MAE was measured for validation, and Accuracy was measured when the tolerance was set to 5 years.

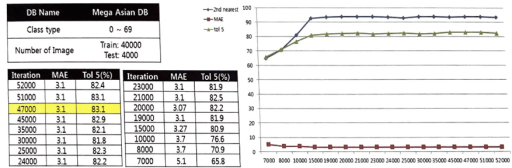

Figure 4. Results of applying Comparative CNN method to Mega Asian DB

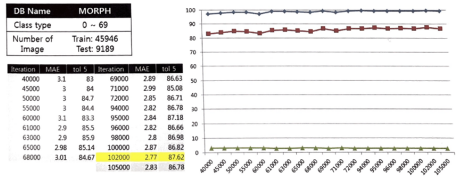

Figure 5. Results of applying Comparative CNN method to MORPH DB

The performance is compared with other techniques and shown in Figure 6. Compared to the Comparative Deep Learning referenced in this paper, the performance improved, as well as the performance improvement in age recognition when compared with the currently best-performing Ranking CNN.

| Method | DB | MAE | Tolerance 6 |
|---|---|---|---|
| Comparative CNN | Mega Asian | 3.1 | 87.4% |
|  | MORPH | 2.77 | 90.2% |
| Ranking CNN (CVPR 2017) | MORPH | 2.96 | 89.9% |
| DEX (VGG net) | Train: IMDB Test: LAP | 3.2 |  |
| Inception | Mega Asian | 3.9 | 74% |
| MS API |  | 6.5 |  |
|  | Mega Asian | 4.0 | 79% |
|  | MORPH | 6.5 |  |
| Comparative Deep Learning | MORPH | 3.74 |  |

Figure 6. Comparison of Performance of Age Estimation Techniques

### 5.2 Results of Sex Classification and Age Estimation

**Gender and Age Estimation Method using Comparative CNN-based Multi-task Learning**

Figure 7 shows the experimental results of applying the technique to the MORPH DB.

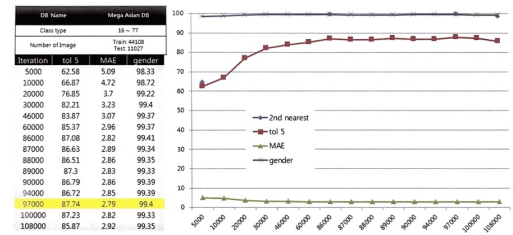

Figure 7. Results of applying gender and age estimation method using Comparative CNN to MORPH

## 6. Conclusion

In this study, gender classification and age estimation were performed using facial images for faces. In addition, to examine algorithms in the field of face recognition using existing machine learning techniques, EigenFace and FisherFace were tested. When using the machine learning technique, the limitations in the classification of lighting changes, facial makeup, and people of color when acquiring images were identified and analyzed.

Looking at the results, it was confirmed that gender classification is a simpler task than age estimation, and a multi-task research method was proposed for gender classification and age estimation. Age estimation is a very difficult problem because the biological age of the face and the age of the actual object are different. The estimation method using Comparative CNN showed significant performance in gender classification and age estimation and was more robust to environmental changes than the existing machine learning method.

If the technology is used to quickly find faces and make them into a database, it is expected that they can be used in various fields such as personal authentication or marketing.

## Acknowledgements

This work was partially supported by ETRI (Electronics and Technology Research Institute) and KyungPook National University Research Fund.